\documentclass[letterpaper, 10 pt, conference]{ieeeconf}  %

\IEEEoverridecommandlockouts                              %

\usepackage{graphicx}
\usepackage[caption=false, font=footnotesize]{subfig}
\usepackage{amsmath} %
\usepackage{amssymb}  %
\usepackage{booktabs}
\usepackage{siunitx}
\usepackage{multirow}
\usepackage{tabularx}
\usepackage[export]{adjustbox}
\usepackage{overpic}
\usepackage{color, colortbl}
\usepackage[usenames,dvipsnames]{xcolor}
\usepackage{xspace}
\usepackage{tikz}
\usepackage[urlcolor=black]{hyperref}

\definecolor{Gray}{gray}{0.9}

\newcommand\resfigwidth{.245}

\usepackage[capitalize]{cleveref}
\crefname{section}{Sec.}{Secs.}
\Crefname{section}{Section}{Sections}
\Crefname{table}{Table}{Tables}
\crefname{table}{Table}{Tables}

\def\eg{\emph{e.g. }} 
\def\ie{\emph{i.e. }}

\def\etal{\emph{et al. }}

\newcommand\copyrighttext{\footnotesize \textcopyright~2023 IEEE. Personal use of this material is permitted. Permission from IEEE must be obtained for all other uses, in any current or future media, including reprinting/republishing this material for advertising or promotional purposes, creating new collective works, for resale or redistribution to servers or lists, or reuse of any copyrighted component of this work in other works.
}

\newcommand\copyrightnotice{%
	\begin{tikzpicture}[remember picture,overlay]
		\node[anchor=south,xshift=0pt,yshift=14pt] at (current page.south) {\fbox{\parbox{\dimexpr\textwidth-\fboxsep-\fboxrule\relax}{\copyrighttext}}};
	\end{tikzpicture}%
}

\title{Joint Out-of-Distribution Detection and Uncertainty Estimation for Trajectory Prediction}

\author{Julian Wiederer$^{1, 2}$, Julian Schmidt$^{1, 3}$, Ulrich Kressel$^{1}$, Klaus Dietmayer$^{3}$ and Vasileios Belagiannis$^{2}$%
\thanks{$^{1}$Julian Wiederer, Julian Schmidt and Ulrich Kressel are with Mercedes-Benz Group AG, 70546 Stuttgart,~Germany. 
{\tt\small \{julian.wiederer, julian.sj.schmidt, ulrich.kressel\}@mercedes-benz.com}}%
\thanks{$^{2}$Julian Wiederer and Vasileios Belagiannis are with the Department of Multi-media Communication and Signal Processing, Friedrich-Alexander-Universität, 91058 Erlangen,~Germany. {\tt\small vasileios.belagiannis@fau.de}}  
\thanks{Julian Schmidt and Klaus Dietmayer are with the Institute of Measurement, Control and Microtechnology, University Ulm, 89081 Ulm,~Germany.
{\tt\small klaus.dietmayer@uni-ulm.de}}%
\thanks{$^{4}$ project page: \url{https://github.com/againerju/joodu}}
}

\begin{document}

\maketitle
\thispagestyle{empty}
\pagestyle{empty}

\begin{abstract}

Despite the significant research efforts on trajectory prediction for automated driving, limited work exists on assessing the prediction reliability.
To address this limitation we propose an approach that covers two sources of error, namely novel situations with out-of-distribution (OOD) detection and the complexity in in-distribution (ID) situations with uncertainty estimation.
We introduce two modules next to an encoder-decoder network for trajectory prediction.
Firstly, a Gaussian mixture model learns the probability density function of the ID encoder features during training, and then it is used to detect the OOD samples in regions of the feature space with low likelihood. Secondly, an error regression network is applied to the encoder, which learns to estimate the trajectory prediction error in supervised training. 
During inference, the estimated prediction error is used as the uncertainty.
In our experiments, the combination of both modules outperforms the prior work in OOD detection and uncertainty estimation, on the Shifts robust trajectory prediction dataset by \SI{2.8}{\%} and \SI{10.1}{\%}, respectively. The code is publicly available$^4$.

\end{abstract}

\copyrightnotice

\section{Introduction}
\label{sec:intro}

Reliable trajectory prediction is critical for the safe motion planning of an automated vehicle. 
However, trajectory prediction failures inevitably occur, for example in out-of-distribution scenarios (OOD)~\cite{malinin2021shiftsdataset, Bahari_2022_CVPR} or even uncertain in-distribution (ID) situations~\cite{ivanovic2022propagating}.   
Yet, the current trajectory prediction approaches~\cite{zhou2022hivt, liang2020learning, salzmann2020trajectron++, varadarajan2022multipath++} do not deliver measures of reliability for the model predictions.

\begin{figure}[t]
    \centering
    \includegraphics[width=\linewidth, trim=0.0cm .1cm 0cm 0.0cm, clip]{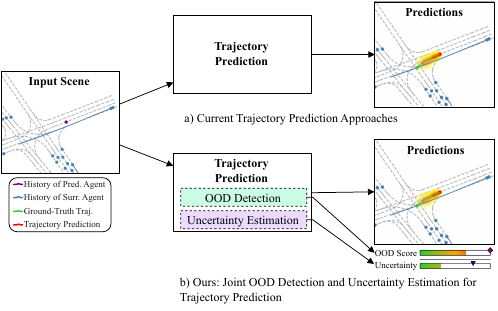}
    \caption{Comparison of (a) current approaches for trajectory prediction and (b) our approach. We introduce two modules for trajectory prediction reliability assessment, namely the out-of-distribution (OOD) detection and the uncertainty estimation. The exemplary scene shows an OOD scenario, resulting in trajectory predictions with high errors. In this case, our introduced module for OOD detection outputs a high OOD score, meaning that the traffic scenario is novel w.r.t. the training data. 
    More qualitative results, also regarding the effectiveness of the uncertainty estimation, are given in \cref{fig:qualitative_results}.
    }
    \label{fig:u_ood_teaser}
\end{figure}

The prior work assesses the reliability of trajectory prediction with OOD detection~\cite{wiederer2022abenchmark, chakraborty2023structural} and uncertainty estimation~\cite{pustynnikov2021estimating, postnikov2021transformer, Gilles2022}. 
OOD detection distinguishes between ID and OOD scenarios, while uncertainty estimation determines the certainty of the prediction regardless of whether it is ID or OOD. 
Both tasks are necessary for safe automated driving.
For instance, a situation with low uncertainty and a high OOD score needs to be treated carefully because it is novel w.r.t. to the training data. In another ID case, high uncertainty also needs to be taken into consideration because the predictions may be unreliable. 
Although both tasks are addressed independently in the literature, a joint formulation is still missing. The only attempt to jointly formulate both tasks is based on model ensembles~\cite{malinin2021shiftsdataset} using Bootstrapped Ensembles~\cite{lakshminarayanan2017simple} or Dropout Ensembles~\cite{gal2016dropout}. Both approaches compute the uncertainty from the output of multiple forward passes during inference, which is not desirable for real-time systems due to the computational cost. 
In contrast to the prior work, we refrain from ensembles and address the aforementioned limitations.

We propose a model-agnostic approach to expand an existing encoder-decoder network for trajectory prediction by joint OOD detection and uncertainty estimation.
Given a trajectory prediction encoder, we introduce two modules to output the OOD score and the uncertainty, respectively.
Our novelty lies in the joint formulation, as well as in our OOD detection approach.
Firstly, inspired by approaches for OOD detection in image classification~\cite{lee2018simple, Ahuja2019probabilistic}, we leverage generative modeling on the encoder features. 
A Gaussian mixture model on the latent representation learns the probability density function of the ID features during training and detects OOD samples in regions of the feature space with low likelihood of this density function.
Secondly, we add an error regression network~\cite{postnikov2021transformer} on the scene encoder for  uncertainty estimation and train it with a supervised regression loss to estimate the prediction error. 
During inference, the estimated error is used as uncertainty.
\cref{fig:u_ood_teaser} illustrates our approach, including the differences to existing trajectory prediction models.
In the experiments on the Shifts dataset~\cite{malinin2021shiftsdataset}, our method outperforms the prior work by a large margin. Therefore, we adopt an existing trajectory prediction approach~\cite{zhou2022hivt} to the new dataset and apply our modules to measure the reliability of the predictions.

In summary, we make the following contributions:
\begin{itemize}
    \item We present a joint model for trajectory prediction with OOD detection and uncertainty estimation, where we add two modules to a shared scene encoder.
    \item We propose a Gaussian mixture model on the latent representation to detect the OOD samples in regions with a low likelihood of the encoder feature space.
    \item We outperform the prior work in both tasks, OOD detection and uncertainty estimation, on the Shifts dataset~\cite{malinin2021shiftsdataset} by a large margin. 
\end{itemize}

\section{Related Work}
\label{sec:related_work}
In the following, we discuss the existing methods for trajectory prediction with a focus on OOD detection and uncertainty estimation.
\noindent\paragraph{Multi-Modal Trajectory Prediction}
Multi-modal trajectory prediction approaches try to cover all possible futures by predicting a set of trajectories~\cite{ zhou2022hivt, liang2020learning, salzmann2020trajectron++, varadarajan2022multipath++, Chai2019MultiPathMP, strohbeck2020multiple} instead of only a single outcome. In recent studies, the encoder-decoder is the default architecture for this task, where the observed scene is encoded into a latent feature space and decoded from the latent space into future motion~\cite{zhou2022hivt, liang2020learning, salzmann2020trajectron++, varadarajan2022multipath++, Chai2019MultiPathMP, gao2020vectornet, schmidt2022crat}. 
While these methods achieve impressive results on the trajectory prediction benchmarks~\cite{chang2019argoverse}, they tend to fail in scenes that are very different from the training data~\cite{Bahari_2022_CVPR}. In addition, in some cases, errors in situations that are similar to the training data can also not be prevented~\cite{malinin2021shiftsdataset}. Examples are highly complex scenarios with many possible outcomes.
In practice, the current methods would have to rely on their prediction in all situations. 
Instead of unconditionally trusting the predictions, we propose two modules to assess the reliability of the prediction, namely the OOD detection and the uncertainty estimation. Both modules are applied to the scene encoder of an existing prediction model.
\noindent\paragraph{Out-of-Distribution Detection for Trajectory Prediction}
The goal of OOD detection is to discriminate the ID from OOD samples. 
While the main research is in the field of computer vision~\cite{hsu2020generalized, hornauer2023heatmap, ren2019likelihood}, in recent years, OOD detection has shown applications in robotics. For example, OOD detection approaches help robots to navigate through novel environments~\cite{Richter2017SafeVN}, drones to avoid collisions in cases other agents show abnormal flying behavior~\cite{sindhwani2020unsupervised} or detect abnormal driving behavior like ghost drivers~\cite{chakraborty2023structural, wiederer22anomaly}.
Likewise, OOD detection can help to detect novel scenarios in trajectory prediction. 
Recently, this task has been addressed by Malinin~\etal~\cite{malinin2021shiftsdataset}. 
They propose to detect the OOD scenarios by aggregating the confidences of multiple prediction models using deep ensembles.
However, ensembles require multiple prediction models during training and inference, which has limited application in real-time systems like automated driving.
In this work, we propose an efficient method attached to the shared scene encoder.
Our approach is closer to the feature-based OOD detection approaches~\cite{lee2018simple, Ahuja2019probabilistic}. 
In particular, we present a Gaussian mixture model on the latent representation for modeling the probability density function of the neural network features and detecting the OOD scenarios in regions of the feature space with low likelihood.  
\noindent\paragraph{Uncertainty for Trajectory Prediction}
While OOD detection helps to assess the prediction quality in cases the test distribution is different from the training distribution, prediction errors in ID scenarios may inevitably occur. In literature, these errors have been tackled by estimating the uncertainty~\cite{malinin2021shiftsdataset}. For example, in deep ensembles, the alignment between the outputs of a set of models is used as uncertainty~\cite{malinin2021shiftsdataset}. Gilles \etal avoid computationally demanding deep ensembles and compute the uncertainty as the integral of the prediction heatmap from a single model~\cite{Gilles2022}. The idea of spectral-normalized Gaussian processes (SNGP) has also been applied to the task of uncertainty estimation~\cite{pustynnikov2021estimating}. The SNGP is directly applied to the scene encoder and the uncertainty is measured by the predicted Gaussian process variance. Similarly, our uncertainty module is applied to the latent features of the scene encoder, but we use error regression to estimate the uncertainty similar to~\cite{postnikov2021transformer}. Our error regression network is trained to approximate the trajectory prediction error. We demonstrate the effectiveness of the approach in comparison with the prior methods on the Shifts dataset~\cite{malinin2021shiftsdataset}. Furthermore, we show that the error regression can estimate the prediction error on ID samples while the Gaussian mixture model can detect the OOD samples.

\section{Method}
\label{sec:method}

Consider the traffic scene represented by a set of $N$ agents and the scene context $\mathcal{I}$, commonly provided by an HD-map. 
The number of agents $N$ can vary between different scenes. 
Each agent $i \in \{1, ..., N\}$ is described by its sequence of states $\mathbf{x}_i = \{\mathbf{s}_i^t\}_{t=-T_h+1}^0$ observed over $T_h$ historical time steps, where $\mathbf{s}_i^t \in \mathbb{R}^d$ contains the $d$ agent states in time step $t$. 
The dynamic scene is summarized in $\mathbf{X} = \{\mathbf{x}_i\}_{i=1}^N$. 
In addition, each agent is assigned the OOD label $\alpha_i \in \{0, 1\}$, indicating if the scene is ID, $\alpha=0$, or OOD, $\alpha=1$.
The ground-truth future trajectory of agent $i$ for the next $T_f$ time steps is denoted as $\mathbf{y}_i = \{(x_i^t, y_i^t)\}_{t=1}^{T_f}$, where $(x_i^t, y_i^t)$ is the agent location in x- and y-coordinates at time step $t$.

Our first goal is to predict the conditional distribution $p(\mathbf{y}_i| \mathbf{X}, \mathcal{I})$ over the future trajectory of an agent $i$ in the scene given the dynamic $\mathbf{X}$ and the static scene context $\mathcal{I}$. Second, we propose to additionally compute an OOD score $\hat{\alpha}_i$, which is small for ID, i.e. $\alpha_i=0$, and large for OOD samples, i.e. $\alpha_i = 1$. 
Thrid, we estimate, at the same time, the uncertainty $\hat{e}_i$ that approximates the true prediction error $e_i$. For that reason, we assume the prediction error $e_i = E(\mathbf{y}_i, p(\mathbf{y}_i| \mathbf{X}, \mathcal{I}))$ between the ground-truth trajectory $\mathbf{y}_i$ and the predicted distribution $p(\mathbf{y}_i| \mathbf{X}, \mathcal{I})$ is measured by the error measurement $E(\cdot, \cdot)$.
We denote the output set for agent $i$ as $\mathbf{Y}_i = \{p(\mathbf{y}_i| \mathbf{X}, \mathcal{I}), \hat{\alpha}_i, \hat{e}_i\}$.

\begin{figure}[t]
    \centering
    \includegraphics[width=\linewidth, trim=0.0cm 0.0cm 0cm 0cm, clip]{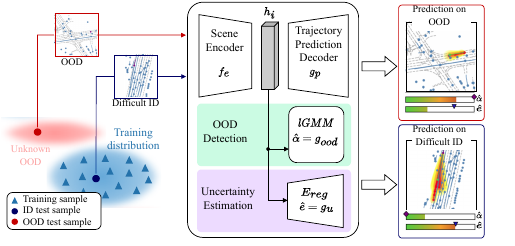}
    \caption{\textbf{Method overview.} In our approach, two modules assess the reliability of a trajectory prediction decoder $g_p$. The OOD detection, realized as a latent Gaussian mixture model \emph{lGMM}, predicts high OOD scores $\hat{\alpha}$ for novel scenarios and the uncertainty estimation, a neural network $E_{reg}$, estimates the true prediction error by generating uncertainty $\hat{e}$. Both modules are applied to the latent feature space of a shared scene encoder $f_e$. We show the inference on two scenes, an OOD scenario (red) and a difficult ID scenario (blue).}
    \label{fig:method}
\end{figure}

\subsection{OOD Detection and Uncertainty Estimation}
\cref{fig:method} gives an overview of our approach. Next, we describe the trajectory prediction network composed of the scene encoder $f_e$ (\cref{sec:scene_encoding}) and the trajectory prediction decoder $g_p$ (\cref{sec:pred_decoder}). 
The scene encoder computes the latent representation $\mathbf{h}_i = f_e(\mathbf{X}, \mathcal{I})$ for each agent $i$ by incorporating social interactions between agents in $\mathbf{X}$ and the scene context $\mathcal{I}$. Subsequently, the trajectory prediction decoder outputs the probability distribution $p(\mathbf{y}_i| \{\mathbf{h}_i\}_{i=1}^N) = g_p(\{\mathbf{h}_i\}_{i=1}^N)$ over the future states $\mathbf{y}_i$ of agent $i$ given all agent encodings $\{{\mathbf{h}}_i\}_{i=1}^N$. To jointly predict the trajectory, OOD score and the uncertainty, we introduce two modules to the scene encoder, namely the OOD detection and the uncertainty estimation. The OOD detection, denoted as $g_{ood}$, predicts a scalar-valued score $\hat{\alpha}_i = g_{ood}(\mathbf{h}_i)$ given the latent feature vector $\mathbf{h}_i$ quantifying if the scene is rather an ID or OOD (\cref{sec:ood_detection}). At the same time, the uncertainty estimation, denoted as $g_u$, estimates the uncertainty in the predicted trajectories $\hat{e}_i = g_u(\mathbf{h}_i)$ from the same features (\cref{sec:pred_u}). The scene encoder is shared across the trajectory prediction decoder and both modules. 
Both, the trajectory prediction decoder and the uncertainty estimation are represented by deep neural networks. For the OOD detection module, we present a Gaussian mixture model on the latent representation, which can detect the outliers in low-density regions of its distribution.

The model is trained in two stages (\cref{sec:training}). 
First, we train the trajectory prediction network and then optimize the two additional modules given the fixed scene encoder. 
We end up with a trajectory predictor and two modules to assess the reliability of the predictions resulting from this predictor. Although we show results with the proposed trajectory prediction model, both modules can be easily adapted to any scene encoder for OOD detection and uncertainty estimation, which outputs a feature vector $\mathbf{h}_i$ per agent.

\subsection{Scene Encoder}
\label{sec:scene_encoding}
Lately, vectorization-based scene encoders have been getting a lot of attention~\cite{ zhou2022hivt, liang2020learning, schmidt2023scene}. Therefore, we use the local encoder from HiVT as the scene encoder $f_e$ and follow the pre-processing as proposed by Zhou~\etal~\cite{zhou2022hivt}. 
Inputs to the scene encoder are the observed states of all agents $\mathbf{X} = \{\mathbf{x}_i\}_{i=1}^N$ and the contextual information $\mathcal{I}$ provided by a set of $L$ lane vectors $\mathbf{V} = \left\{\mathbf{v}_l\right\}_{l=1}^L$ with $\mathbf{v}_l \in \mathbb{R}^o$ containing $o$ lane features. 
At first, we generate a translation invariant scene representation. 
Therefore, the scene elements including the past state vectors and the lane segments from the HD-map are transformed into a vector representation and augmented with relative position vectors between elements to preserve distance information. Then, for each agent $i$ the encoder 
extracts the spatio-temporal features $\mathbf{h}_{i} = f_e(\mathbf{x}_{n\in\mathcal{N}_i}, \mathbf{v}_{m\in\mathcal{N}_i})$
from the observed state sequence of the agents $\mathbf{x}_{n\in \mathcal{N}_i} \subseteq \mathbf{X}$ and the lanes $\mathbf{v}_{m\in \mathcal{N}_i} \subseteq \mathbf{V}$ in the local neighborhood $\mathcal{N}_i$, determined by a circle with radius $r$ around the corresponding agent. 
The encoder outputs the set of all agent feature vectors $\{\mathbf{h}_i\}_{i=1}^N$.

\subsection{Multi-modal Trajectory Prediction Decoder} 
\label{sec:pred_decoder}
To cover multiple trajectory modes, we make the assumption that the target distribution $p(\mathbf{y}_i| \{\mathbf{h}_i\}_{i=1}^N)$ follows a mixture density distribution~\cite{bishop1994mixture}.
Each density component $k \in \{1, ...,K\}$ represents one of $K$ possible future trajectories and is described by a sequence of independent bi-variate Gaussian distributions, with the joint distribution defined as $\prod_{t=1}^{T_f} \mathcal{N}(\boldsymbol\mu_{i,k}^t, \boldsymbol\Sigma_{i,k}^t)$ over the $T_f$ future time steps with the center location $\boldsymbol\mu_{i,k}^t \in \mathbb{R}^2$ and the covariance $\boldsymbol\Sigma_{i,k}^t\in \mathbb{R}^{2\times 2}$ in time step $t$. 
The density components are combined as a weighted-sum using the mixing coefficients $\pi_{i, k}$ from the categorical distribution $q(k) = \pi_{i,k}$, with $\sum_{k=1}^K \pi_{i,k} = 1$. 
We summarize the set of covariances, center locations and mixing coefficients for agent $i$ with $\boldsymbol\Sigma_i$, $\boldsymbol\mu_i$ and $\boldsymbol\pi_i$.
The resulting density over the future trajectory is defined as
\begin{equation}
    p(\mathbf{y}_i| \pi_i, \boldsymbol\mu_i, \boldsymbol\Sigma_i) = \sum_{k=1}^K \pi_{i, k} \prod_{t=1}^{T_f} \mathcal{N}(\boldsymbol\mu_{i,k}^t, \boldsymbol\Sigma_{i,k}^t).
\end{equation}
We assume the  diagonal covariance $\boldsymbol\Sigma_{i,k}^t = (\sigma_{i,k}^t)^2\mathbf{I}$, with the identity matrix $\mathbf{I} \in \mathbb{R}^{2\times 2}$ and standard deviation $\sigma_{i, k}^t$; and denote $\hat{\mathbf{y}}_{i,k} = \{\boldsymbol{\mu}_{i,k}^t\}_{t=1}^{T_f}$ the mean trajectory of mode $k$ and agent $i$.
The distribution parameters $\{\boldsymbol\Sigma_i, \boldsymbol\mu_i, \boldsymbol\pi_i\}_{i=1}^N = g_p(\{\mathbf{h}_i\}_{i=1}^N)$ for all agents are predicted by the trajectory prediction decoder $g_p$ given the encoding features $\{\mathbf{h}_i\}_{i=1}^N$. The decoder is composed of the global message passing network and the aggregation network as proposed by~\cite{zhou2022hivt}, and a separate multi-layer perceptron (MLP) for each distribution parameter. For the mixing coefficients, we predict the unnormalized coefficients $\tilde{\pi}_{i,k}$ first, and use the softmax function to convert them to probabilities $\pi_{i,k}$~\cite{bishop1994mixture}. We denote the resulting trajectory prediction model \emph{HiVT$^*$}.

\subsection{Out-of-Distribution Detection}
\label{sec:ood_detection}
Since OOD scenarios are rare and in most cases inaccessible during training, we describe OOD detection as one-class classification problem, i.e. only ID samples are available during training. We propose an OOD detector on the latent representation space of the scene encoder to discriminate OOD from ID samples. To this end, we estimate the parameters of a parametric probability distribution representing the ID from the training feature vectors and identify the OOD samples in regions with low density during testing. Our OOD detection module $g_{ood}$ is based on the assumption that the probability distribution of the latent features $p(\mathbf{h}_i)$ follows a mixture of multivariate Gaussian distributions~\cite{lee2018simple, Ahuja2019probabilistic}. Therefore we define the Gaussian mixture model $q(\mathbf{h}_i) = \sum_{i=1}^C \phi_c \mathcal{N}(\mathbf{h}_i|\boldsymbol\mu_c, \boldsymbol\Sigma_c)$, where $\phi_c$ is the mixing coefficient, $\boldsymbol\mu_c$ is the mean and $\boldsymbol\Sigma_c$ is the covariance matrix of mixture component $c = \{1, ..., C\}$. 
We denote the OOD detector as latent GMM (\emph{lGMM}) throughout the experiments. 
During inference, the \emph{lGMM} outputs the OOD score 

\begin{equation}
    \hat{\alpha}_i = -\log q(\mathbf{h}_i) = -\log \left( \sum_{i=1}^C \phi_c \mathcal{N}(\mathbf{h}_i|\boldsymbol\mu_c, \boldsymbol\Sigma_c) \right)
\end{equation}

as the negative log-likelihood under the Gaussian mixture distribution $q(\mathbf{h}_i)$, where an ID scenario has a low and an OOD scenario has a high negative log-likelihood.

\subsection{Uncertainty Estimation}
\label{sec:pred_u}
To accurately detect prediction errors in ID scenarios, we introduce an uncertainty estimation network, denoted as $g_u$, which is a small MLP applied to the scene encoder. 
We formulate the problem of uncertainty estimation as a regression task, similar to~\cite{postnikov2021transformer}, and train the neural network to predict the true trajectory prediction error $e_i$ for the agent $i$, given the encoder feature vector $\mathbf{h}_i$. 
During inference, the error regression network, we denote as $E_{reg}$ throughout the experiments, outputs the uncertainty as  
\begin{equation}    
    \hat{e}_i = g_u(\mathbf{h}_i).
\end{equation}

\subsection{Two-phase Model Training}
\label{sec:training}
Our training process is divided into two phases. First, we learn the parameters of the scene encoder $f_e$ and the trajectory prediction decoder $g_p$ by optimizing the prediction loss $\mathcal{L}_p$. The prediction loss $\mathcal{L}_{p, i}$ for the agent $i$ is defined as the negative log-likelihood on the mixture of Gaussian distributions
\begin{equation}
\begin{split}
    \mathcal{L}_{p, i} = & -\log \left [ \sum_{k=1}^{K} \pi_{i,k} \prod_{t=1}^{T_f} \mathcal{N}\left(\mathbf{y}_i^{t} | \hat{\mathbf{y}}_{i, k}^t, \boldsymbol\Sigma_{i,k}^t\right) \right ]
\end{split}
\end{equation}
with the locations of the ground truth trajectory $\mathbf{y}_i^t$, the predicted locations of the $k$ mixture modes $\hat{\mathbf{y}}_{i, k}^t = \boldsymbol\mu_{i, k}^t$ and the corresponding covariance matrices $\boldsymbol\Sigma_{i,k}^t$. 

Once the trajectory prediction model is trained, the OOD detection and uncertainty estimation modules are optimized. 
During this stage, in order to avoid impacting the trajectory prediction, both the weights of the scene encoder and the trajectory prediction decoder are fixed.
The parameters $\{\phi_c, \boldsymbol\mu_c, \boldsymbol\Sigma_c\}_{c=1}^C$ of the \emph{lGMM} are estimated using the EM-Algorithm~\cite{bishop2006pattern}. The $E_{reg}$ is trained in a supervised manner with the error regression loss $\mathcal{L}_{u,i}$ defined as the mean-squared-error 
\begin{equation}
    \mathcal{L}_{u, i} = \lVert (e_i - \hat{e}_{i}) \rVert_2
\end{equation}
between the estimated error $\hat{e}_{i}$ and the true prediction error $e_i$. The prediction error $e_i$ can be set to any error measurement $E(\cdot, \cdot)$, for example the prediction loss $e_i = \mathcal{L}_{p, i}$ or one of the evaluation metrics as explained in the experiments \cref{sec:exp_setup}. Since the training is decoupled from the training of the trajectory prediction method, our reliability modules can be considered as post-hoc methods. They can be easily added to an existing trajectory prediction model without the need for expensive re-training.

\section{Experiments}
\label{sec:exp}
We first describe the experimental setup, consisting of the Shifts dataset~\cite{malinin2021shiftsdataset}, the evaluation protocol and our implementation. Secondly, our approach is compared with the baselines for trajectory prediction, OOD detection and uncertainty estimation. Finally, we visualize qualitative results and analyze the runtime.

\subsection{Experimental Setup}
\label{sec:exp_setup}

\textbf{Dataset.} We evaluate our method on the Shifts dataset~\cite{malinin2021shiftsdataset}. Shifts is unique because it is the only dataset available for OOD detection and uncertainty estimation in trajectory prediction. It consists of an ID training set (\emph{train}) with \SI{388406}{} sequences recorded during drives in Moscow without precipitation. In contrast to training, \SI{9569}{} of the \SI{36605}{} validation sequences (\emph{dev}) and \SI{9939}{} of the \SI{36804}{} test sequences (\emph{eval}) are affected by distribution shifts, denoted as OOD samples. Therefore, we can form three \emph{dev} and three \emph{eval} sets denoted as \emph{ID}, \emph{OOD} and \emph{Full}. The OOD samples are recorded in adverse weather conditions like rain or snow and cities like Ann-Arbor or Tel Aviv. Each scene contains the states of all dynamic agents, including pedestrians and vehicles, in a 2D bird's-eye-view coordinate system recorded at \SI{5}{Hz}. Pedestrians and vehicles are described by the position vector and the velocity vector in each time step, vehicles with additional acceleration and yaw angle. HD-maps are provided for all locations. The $10$ second recordings are divided into $5$ seconds observation and $5$ seconds prediction horizon.

\textbf{Evaluation Protocol.}
Our evaluation is three-fold. First, we evaluate trajectory prediction and then OOD detection and uncertainty estimation. The agent index $i$ is discarded in the definition of the metrics. 

To measure the quality of the predicted trajectories, we use the minimum average displacement error (\emph{minADE}) and the minimum final displacement error (\emph{minFDE})~\cite{zhou2022hivt}. In addition, we consider the more practical weighted ADE, denoted as $wADE(\mathbf{y}, \hat{\mathbf{y}}) = \sum_{k = 1}^{K} \pi_k \cdot ADE(\mathbf{y}, \hat{\mathbf{y}}_k)$~\cite{malinin2021shiftsdataset}, where we use the mixture coefficients $\pi_k$ to average the ADE over the $K$ modes, which we denote as our main metric. Analogously, we compute the \emph{wFDE}. While highly intuitive, the metrics have two limitations: They do not include the predicted distribution and suffer from mode-collapse as derived by Malinin~\etal~\cite{malinin2021shiftsdataset}. 
We address both limitations by providing the negative log-likelihood score as introduced in~\cite{ivanovic2019trajectron}. The \emph{NLL} evaluates the likelihood of the ground truth trajectory under the predicted Gaussian mixture distribution.
All metrics are calculated for the \emph{ID}, the \emph{OOD} and the \emph{Full} \emph{eval} set.

We use the standard metric to evaluate the OOD detection~\cite{DBLP:journals/corr/HendrycksG16c}, 
namely the area under the receiver operating characteristic curve (\emph{AUROC}). 
The ROC curve plots the true positive rate against the false positive rate at various thresholds of the predicted OOD score. 
The area under the ROC curve evaluates if the OOD scores can detect the OOD samples. 
An optimal classifier achieves \SI{100}{\%} \emph{AUROC}, a random classifier \SI{50}{\%} \emph{AUROC}. 
The uncertainty estimation is evaluated by computing the area under the retention curve (\emph{R-AUC})~\cite{malinin2021shiftsdataset}. 
Retention curves measure the agreement between the uncertainty $\hat{e}$ and the prediction error $e$, which can be computed by any error metric, \eg $e = wADE$. 
First, the list of all prediction errors $e$ on the \emph{eval} set is sorted with descending uncertainty $\hat{e}$. 
Then iteratively, the subset of samples with the highest uncertainty is discarded and the error $e$ is averaged over the remaining samples. 
If the uncertainty properly represents the error, the average error is supposed to shrink with the decreasing retention fraction. The optimal retention curve is obtained by sorting the samples in descending order of the true prediction error $e$, subsequently denoted as \emph{Oracle}. We use the \emph{wADE} as the error metric, \ie $e=wADE$, to compute the \emph{wADE R-AUC}. 

\begin{figure}
    \centering
    \includegraphics[width=0.95\linewidth, trim=0cm 0.1cm 0cm .12cm, clip]{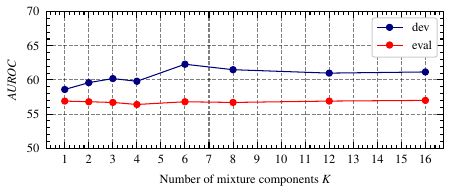}
    \caption{Hyperparameter search. \emph{AUROC} over the number of \emph{lGMM} components $K$ on the \emph{Full} \emph{dev} and \emph{eval} set.}
    \label{fig:ablation_gmm_components}
\end{figure}

\textbf{Implementation Details.} 
The agent states $\mathbf{s}_i^t \in \mathbb{R}^7$ are filled with the vectorized position, the velocity and the acceleration in x- and y-coordinates, as well as a binary flag indicating whether the agent is a vehicle or a pedestrian. We set the acceleration to zero for pedestrians. Each lane vector $\mathbf{v}_l \in \mathbb{R}^{10}$ contains the x- and y-coordinate of a vectorized centerline segment plus a set of context features derived from the HD-map: The speed limit, the lane availability vector, which is derived from the traffic light state, and the lane priority. 
The size of the scene encoder features $\mathbf{h}_i \in \mathbb{R}^{128}$ is set to $128$ and the radius of the receptive field to $r = 50$ meters~\cite{zhou2022hivt}.
The trajectory prediction decoder uses a couple of MLPs to output the distribution parameters: A three-layer MLP for the mixing coefficients $\tilde{\boldsymbol\pi}_{i, k}$ and two two-layer MLPs for the mean $\boldsymbol\mu_{i,k}$ and the variance $\boldsymbol\sigma_{i,k}^2$, respectively. Our motion decoder predicts a set of $K = 5$ trajectory modes as defined in Shifts~\cite{malinin2021shiftsdataset}. The uncertainty decoder $E_{reg}$ is composed of a three-layer MLP to estimate $\hat{e}_{i}$. 

In the first phase of training, the prediction loss $L_p$ is optimized for $64$ epochs with initial learning rate $1\times10^{-4}$ and batch size $48$ on four NVIDIA Tesla V100 GPUs. In the second phase, the weights of the $E_{reg}$ module are learned using the regression loss $L_u$ for $100$ epochs with learning rate $1\times10^{-3}$ and batch size $1024$. We define $wADE(\mathbf{y}_i, \hat{\mathbf{y}}_i)$ as the error function $E(\cdot, \cdot)$ to compute the regression target $e_i = \log(wADE(\mathbf{y}_i, \hat{\mathbf{y}}_i))$ and scale it with the logarithm to reduce the output range. Both training phases use the AdamW optimizer~\cite{Loshchilov2019DecoupledWD} with a cosine annealing learning rate scheduler~\cite{Loshchilov2017SGDR}. 
Simultaneously, the parameters of the $lGMM$ are fit using the EM-Algorithm for a maximum of 100 iterations, after initialization with the k-means algorithm. We choose $K = 6$ mixture components from $K\in\{1, 2, 3, 4, 6, 8, 12, 16\}$ based on the highest \emph{AUROC} on the \emph{dev} set. The hyperparameter search is shown in~\cref{fig:ablation_gmm_components}.

\begin{table*}[ht!]
    \centering
    \caption{Trajectory prediction results on the Shifts \emph{eval} set~\cite{malinin2021shiftsdataset}.}
\vspace{0.1in}
\resizebox{1\textwidth}{!}{%
    \begin{tabular}{@{}l|cc|c|ccc|ccc|ccc|ccc|ccc}
        \toprule
        \multirow{2}*{\textbf{Model}} & \multicolumn{2}{c|}{\textbf{Scene Input}} & \multirow{2}*{\textbf{Ensemble}} & \multicolumn{3}{c|}{\textbf{wADE} $\downarrow$} & \multicolumn{3}{c|}{\textbf{minADE} $\downarrow$} & \multicolumn{3}{c}{\textbf{wFDE} $\downarrow$} &  \multicolumn{3}{c|}{ \textbf{minFDE} $\downarrow$} & \multicolumn{3}{c}{\textbf{NLL} $\downarrow$} \\
        
        & rast. & vect. & & ID & OOD & Full & ID & OOD & Full & ID & OOD & Full & ID & OOD & Full & ID & OOD & Full  \\
        
        \midrule
        \rowcolor{Gray}
        BC (K=1)~\cite{malinin2021shiftsdataset} & $\bullet$ & & & 1.104 & 1.407 & 1.164 & 0.829 & 1.084 & 0.880  & 2.394 & 3.197 & 2.555 & 1.733 & 2.420 & 1.870 & 106.15 & 144.77 & 113.88  \\
        
        BC (K=5)~\cite{malinin2021shiftsdataset} & $\bullet$ & & $\bullet$ & \textbf{1.028} & \underline{1.299} & \underline{1.082} & 0.777 & 1.014 & 0.824 & \textbf{2.238} & \underline{2.957} & \textbf{2.382}  & 1.636 & 2.278 & 1.765 & 103.70 & 140.95 & 111.15 \\

        \rowcolor{Gray}
        DIM (K=1)~\cite{malinin2021shiftsdataset} & $\bullet$ & & & 1.551 & 1.883 & 1.618 & 0.759 & 0.942 & 0.796 & 3.536 & 4.376 & 3.704 & 1.511 & \underline{1.983} & 1.605 & \underline{96.45} & \underline{121.95} & 101.55 \\
        
        DIM (K=5)~\cite{malinin2021shiftsdataset} & $\bullet$ & & $\bullet$ & 1.424 & 1.754 & 1.490 & \underline{0.728} & \underline{0.918} & 0.766 & 3.256 & 4.093 & 3.424 & \underline{1.493} & 2.000 & 1.595 & 97.14 & 124.80 & 102.68 \\

        \rowcolor{Gray}
        VNT~\cite{pustynnikov2021estimating} & & $\bullet$ &  & - & - & 1.326 & - & - & \underline{0.495} & - & - & 3.158 & - & - & \underline{0.936} & - & - & 61.55 \\
        
        ViT~\cite{postnikov2021transformer} & $\bullet$ & & & - &  -  & 1.850 & - &  -  & 0.526  & - &  - & 4.433 & - &  - & 1.016 & - & - & \underline{61.63} \\
        
        \midrule

        \rowcolor{Gray}
        HiVT$^*$ & & $\bullet$ & & \underline{1.031} & \textbf{1.234} & \textbf{1.072} & \textbf{0.394} & \textbf{0.473} & \textbf{0.410} & \underline{2.367} & \textbf{2.933} & \underline{2.480} & \textbf{0.748} & \textbf{0.968} & \textbf{0.792} & \textbf{-20.42} & \textbf{-16.44} & \textbf{-19.62} \\
        \bottomrule
    \end{tabular}
}

    \label{table:res_trajectory_prediction}
\end{table*}

\subsection{Comparison with State-of-the-Art}
The results are split into two parts: The evaluation of trajectory prediction, see \cref{table:res_trajectory_prediction}, and the evaluation of OOD detection and uncertainty estimation, see \cref{table:u_ood_results}. In the tables the highest scores are denoted in \textbf{bold} and the second highest scores are \underline{underlined}. The symbol "-" means, that the results have not been reported. The arrows, $\downarrow$ and $\uparrow$, indicate the direction of better performance.

\textbf{Multi-modal Trajectory Prediction.} We compare our trajectory prediction method \emph{HiVT$^*$} with six state-of-the-art models on the Shifts \emph{eval} set in \cref{table:res_trajectory_prediction}. \emph{BC} and \emph{DIM} use rasterized scene inputs and output a set of trajectories by sampling from a learned uni-modal Gaussian distribution. In addition to the single \emph{BC} and \emph{DIM} models with $K=1$, we show the results of the ensembles with $K=5$~\cite{malinin2021shiftsdataset}. In particular, the single model variants with $K=1$ perform poorly throughout all metrics. Averaging the results over the bootstrapped ensemble helps to boost performance, especially on the \emph{wADE} and \emph{wFDE}. However, ensembles involve multiple forward passes and therefore have limited applicability in real-time systems like automated vehicles. 
\emph{VNT}~\cite{pustynnikov2021estimating} combines a graph-based encoder on a vectorized scene with a transformer-based decoder and \emph{ViT}~\cite{postnikov2021transformer} a vision transformer-based encoder on a rasterized scene with an MLP decoder. Both models predict a Gaussian mixture distribution over future trajectories, but only predict a deterministic mean without a covariance, which is set to the identity matrix  $\boldsymbol\Sigma = \mathbf{I}$ during training. In contrast, we follow a probabilistic approach and predict the standard deviation in addition to the mean locations with the covariance matrix defined as $\boldsymbol\Sigma_{i,k}^t = (\sigma_{i,k}^t)^2\mathbf{I}$. This is advantageous on the \emph{NLL} metric, which particularly evaluates the probabilistic multi-modal predictions. 
Overall our method out-performs the baselines in terms of \emph{minADE}, \emph{minFDE} and \emph{NLL} and shows first and second best results on \emph{wADE} and \emph{wFDE}. Notice, all models show a performance drop on the \emph{OOD} set in comparison to the \emph{ID} set.

\begin{table}
\scriptsize
\centering
\caption{OOD detection and uncertainty estimation results on the Shifts \emph{eval} set~\cite{malinin2021shiftsdataset}. The $\hat{\alpha}$ and $\hat{e}$ columns describe the method to predict the OOD score and the uncertainty, respectively.}
\resizebox{\linewidth}{!}{
\begin{tabular}{@{}l|c|c|c|ccc}

     \multicolumn{3}{c}{} & \multicolumn{1}{|c}{\textbf{OOD Detection} $\hat{\alpha}$} & \multicolumn{3}{|c}{\textbf{Uncertainty Estimation} $\hat{e}$} \\

    \midrule
    
    \textbf{Prediction} & \multirow{2}*{$\hat{\boldsymbol\alpha}$} & \multirow{2}*{$\hat{\mathbf{e}}$} &  \multirow{2}*{\textbf{AUROC (\%)} $\uparrow$} & \multicolumn{3}{c}{\textbf{wADE R-AUC} $\downarrow$} \\
     \textbf{Model} &  &  & & ID & OoD & Full \\
    
    \midrule

    BC (K=1)~\cite{malinin2021shiftsdataset} & \multicolumn{2}{c|}{MA} & 52.8 & - & - & 0.293  \\
    
    \rowcolor{Gray} BC (K=5)~\cite{malinin2021shiftsdataset} & \multicolumn{2}{c|}{MA} & 52.1 & - & - & 0.258 \\
    
    DIM (K=1)~\cite{malinin2021shiftsdataset} & \multicolumn{2}{c|}{MA} & 51.8  & - & - & 0.458 \\

    \rowcolor{Gray}
    DIM (K=5)~\cite{malinin2021shiftsdataset} & \multicolumn{2}{c|}{MA} & 50.9 & - & - & 0.411 \\

    VNT~\cite{pustynnikov2021estimating} & \multicolumn{2}{c|}{SNGP} & 54.0 & - & - & 0.327 \\

    \rowcolor{Gray}
    ViT~\cite{postnikov2021transformer} & \multicolumn{2}{c|}{$E_{reg}$} & 53.1 & - & - & 0.455 \\  

    \midrule
    
    HiVT$^*$ & \multicolumn{2}{c|}{NLL} & 50.3 & \underline{0.232} & \underline{0.293} & \underline{0.244} \\

    \rowcolor{Gray}
    HiVT$^*$ & \multicolumn{2}{c|}{MA} & \underline{54.6} & 0.421 & 0.525 & 0.440 \\

    HiVT$^*$ & \multicolumn{2}{c|}{SNGP} & 52.6 & 0.253 & 0.316 & 0.265 \\

    \rowcolor{Gray}
    HiVT$^*$ & \multicolumn{2}{c|}{$E_{reg}$} & 52.3 & \textbf{0.222} & \textbf{0.272} & \textbf{0.232} \\

    HiVT$^*$ & \multicolumn{2}{c|}{$l$GMM} & \textbf{56.8} & 0.436 & 0.462 & 0.440 \\

    \midrule

    \rowcolor{Gray}
    {HiVT$^*$} & $l$GMM & $E_{reg}$ &{\textbf{56.8}} & {\textbf{0.222}} &{\textbf{0.272}} & {\textbf{0.232}} \\ 
    
\bottomrule
\end{tabular}
}
\label{table:u_ood_results}
\end{table}

\textbf{OOD Detection and Uncertainty Estimation.} \cref{table:u_ood_results} shows the results of our approach, the combination of \emph{lGMM} and $E_{reg},$ in comparison with the prior work on OOD detection and uncertainty estimation, respectively. The results of \emph{VNT} and \emph{ViT} are collected from the Shifts leaderboard~\cite{malinin2021shiftsdataset} on 07/31/2023.
In contrast to our approach, where we address both tasks with different methods, the prior works use the predicted uncertainty $\hat{e}$ as OOD score $\hat{\alpha}$, simultaneously. The uncertainty of the \emph{BC} and \emph{DIM} backbones~\cite{malinin2021shiftsdataset} is computed by model averaging (\emph{MA}), \ie averaging the confidences of all trajectories predicted by the single model for $K=1$ or the ensemble for $K=5$. The ensembles, $K=5$, reach lower \emph{wADE R-AUC} compared to the single models, $K=1$, while the single models are slightly better on \emph{AUROC}. For the \emph{VNT}, the uncertainty is estimated by the predicted variance of a spectral-normalized Gaussian process (\emph{SNGP}), which is applied to the encoder feature space~\cite{pustynnikov2021estimating}. \emph{ViT} uses error regression $E_{reg}$~\cite{postnikov2021transformer} to estimate the uncertainty. 
For a fair comparison and consistent results, we adopt the prior approaches to our trajectory prediction model. We do not consider ensembles due to their limited application in real-time systems, i.e. $K = 1$ in all experiments. 

\begin{figure}
    \centering
    \includegraphics[width=0.95\linewidth, trim=0cm 0.1cm 0cm .12cm, clip]{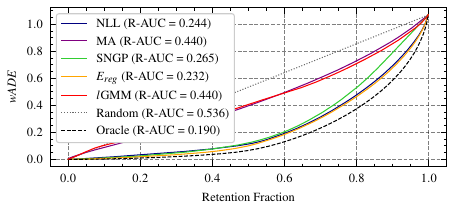}
    \caption{Retention curves of the uncertainty estimation methods. The \emph{wADE} is plotted over the retention fraction. The \emph{Random} estimator reaches 0.536 \emph{wADE R-AUC} and the \emph{Oracle} 0.190 \emph{wADE R-AUC}. The proposed $E_{reg}$ is close to the Oracle with 0.232 \emph{wADE R-AUC}.} 
    \label{fig:retention_curves}
\end{figure}

In addition to \emph{MA}~\cite{malinin2021shiftsdataset}, \emph{SNGP}~\cite{pustynnikov2021estimating} and $E_{reg}$~\cite{postnikov2021transformer}, we present another simple yet effective baseline. To this end, we compute the uncertainty as the negative log-likelihood under the output Gaussian mixture distribution $p(\mathbf{y}_i| \pi_i, \boldsymbol\mu_i, \boldsymbol\Sigma_i)$ and denote the baseline \emph{NLL}. The idea is, that the model is certain about its prediction for small standard deviations and uncertain otherwise. 
\emph{NLL} performs well on uncertainty estimation, but falls behind $E_{reg}$. We compare the retention curves in~\cref{fig:retention_curves}.
From the results it becomes clear, that it is not sufficient to use the same method for OOD detection and uncertainty estimation, since none of the prior approaches performs well on both tasks, simultaneously. Therefore, we propose to learn both tasks jointly with two expert models, namely \emph{lGMM} for OOD detection and $E_{reg}$ for uncertainty estimation. 
On the one hand, \emph{lGMM} is best for OOD detection, because it learns a probability density function over the ID scenarios during training, while on the other hand, $E_{reg}$ outperforms the baselines for uncertainty estimation, due to the supervised training on the \emph{wADE} error. 
Our \emph{lGMM} significantly outperforms all prior approaches by at least \SI{2.8}{\%}, including the ensemble models~\cite{malinin2021shiftsdataset}, on OOD detection.

\begin{figure*}[t!] %

    \centering
    \subfloat{%
        \includegraphics[width=\resfigwidth\linewidth, trim=1.0cm 0.5cm 1.0cm 0.9cm, clip, frame]{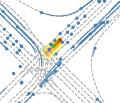}}
    \hfill
    \subfloat{%
        \includegraphics[width=\resfigwidth\linewidth, trim=0.9cm 0.7cm 1.1cm 0.7cm, clip, frame]{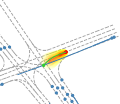}}
    \hfill
    \subfloat{%
        \includegraphics[width=\resfigwidth\linewidth, trim=.6cm 0.7cm 1.4cm 0.7cm, clip, frame]{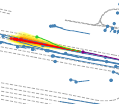}}
    \hfill
    \subfloat{%
        \includegraphics[width=\resfigwidth\linewidth, trim=1cm .5cm 1.0cm .9cm, clip, frame]{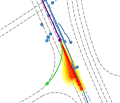}}
    \\
    \subfloat{%
        \begin{overpic}[width=\resfigwidth\linewidth, trim=-1.35cm 0cm 0cm 0cm, clip]{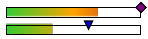}
            \scriptsize
            \put(1, 10){OOD Score $\hat{\alpha}$}
            \put(1, 2){Uncertainty $\hat{e}$}
        \end{overpic}}
    \hfill
    \subfloat{%
        \begin{overpic}[width=\resfigwidth\linewidth, trim=-1.35cm 0cm 0cm 0cm, clip]{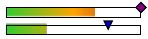}
            \scriptsize
            \put(1, 10){OOD Score $\hat{\alpha}$}
            \put(1, 2){Uncertainty $\hat{e}$}
        \end{overpic}}
    \hfill
    \subfloat{%
        \begin{overpic}[width=\resfigwidth\linewidth, trim=-1.35cm 0cm 0cm 0cm, clip]{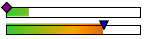}
            \scriptsize
            \put(1, 10){OOD Score $\hat{\alpha}$}
            \put(1, 2){Uncertainty $\hat{e}$}
        \end{overpic}}
    \hfill
    \subfloat{%
        \begin{overpic}[width=\resfigwidth\linewidth, trim=-1.35cm 0cm 0cm 0cm, clip]{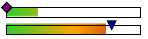}
            \scriptsize
            \put(1, 10){OOD Score $\hat{\alpha}$}
            \put(1, 2){Uncertainty $\hat{e}$}
        \end{overpic}}
    \caption{Qualitative results on the Shifts \emph{eval} set. The lane centerlines are shown in grey, the past trajectory of the target agent in purple and the ground-truth future trajectory in green. The observed trajectories of the other agents are shown in blue. Circular markers indicate the endpoints of trajectories. We illustrate $c=\{1, 2, 3\}$ multiples of the covariance matrices $\boldsymbol\Sigma_{j,k}^t=c\ \sigma_{j,k}^t\mathbf{I}$ as ellipsoids around the mean trajectory $\boldsymbol\mu_{j,k}^t$ (shown as red line), with $c = 1$ in red, $c=2$ in orange and $c=3$ in yellow. The transparency of the ellipsoids $\gamma_k = \pi_k$ is set according to the coefficients of the categorical distribution $\pi_k$. Below the scene, the OOD score $\hat{\alpha}$ is shown as a gradient bar. The purple diamond indicates the binary ground-truth label $\alpha \in \{0, 1\}$ and is placed left for ID and right for OOD scenarios. The second gradient bar shows the uncertainty $\hat{e}$ with the true prediction error $e$ indicated by a blue triangle. Both quantities are normalized on the \emph{dev} set.}
    \label{fig:qualitative_results}
\end{figure*}

\subsection{Results on the Shifts Motion Prediction Challenge}
In the following, we provide the results on the two metrics defined for the Shifts motion prediction challenge~\cite{malinin2021shiftsdataset}. This is for trajectory prediction the corrected NLL defined as $cNLL = NLL - T_f \log (2\pi)$ by subtracting $T_f \log (2\pi)$ from \emph{NLL} to ensure the minimum value is zero~\cite{malinin2021shiftsdataset} and for uncertainty estimation, the area under the retention curve of the \emph{cNLL} denoted as \emph{cNLL R-AUC}. Although used in the challenge, the \emph{cNLL} is less expressive than the \emph{NLL}, since the correction assumes the constant covariance $\boldsymbol\Sigma=\mathbf{I}$, which is clearly not desirable for probabilistic motion prediction where the variance can take on any value. Nevertheless, we train a variant of our model with the variance fixed to one, denoted as \emph{HiVT$^*$ ($\boldsymbol\Sigma_{i,k}^t=\mathbf{I}$}), and optimize both \emph{lGMM} and $E_{reg}$ on the new scene encoder. We compare the results with the prior approaches~\cite{malinin2021shiftsdataset, pustynnikov2021estimating, postnikov2021transformer} in~\cref{table:shifts_challenge}. Our model outperforms the baselines in all metrics with an impressive improvement of \SI{23.4}{\%} on \emph{cNLL} over the second best \emph{ViT}.

\begin{table}
\centering
\caption{Results on the Shifts motion prediction challenge~\cite{malinin2021shiftsdataset}.}
\resizebox{\linewidth}{!}{
\begin{tabular}{@{}l|c|c|ccc|c|c}
     \multicolumn{3}{c}{} & \multicolumn{3}{|c}{\textbf{Trajectory Prediction}} & \multicolumn{1}{|c}{\textbf{OOD} $\hat{\alpha}$} & \multicolumn{1}{|c}{\textbf{Uncertainty} $\hat{e}$} \\

    \midrule

     \multirow{2}*{\textbf{Backbone}} & \multirow{2}*{$\hat{\boldsymbol\alpha}$} & \multirow{2}*{$\hat{\mathbf{e}}$} & \multicolumn{3}{c|}{\textbf{cNLL} $\downarrow$} & \multirow{2}*{\textbf{AUROC (\%)} $\uparrow$} & \textbf{cNLL R-AUC} $\downarrow$  \\
     & & & ID & OoD & Full &  & Full \\
    \midrule
    \rowcolor{Gray}
    BC (K=1)~\cite{malinin2021shiftsdataset} & \multicolumn{2}{c|}{MA} & 60.20 & 98.82 & 67.93 & 52.8 & 12.91 \\
    
    BC (K=5)~\cite{malinin2021shiftsdataset} & \multicolumn{2}{c|}{MA} & 57.75 & 95.00 & 65.20 & 52.1 & 10.57 \\
    
    \rowcolor{Gray}
    DIM (K=1)~\cite{malinin2021shiftsdataset} & \multicolumn{2}{c|}{MA} & \underline{50.50} & \underline{76.00} & 55.60 & 51.8 & 14.32 \\
    
    DIM (K=5)~\cite{malinin2021shiftsdataset} & \multicolumn{2}{c|}{MA} & 51.19 & 78.85 & 56.73 & 50.9 & 15.16 \\

    \rowcolor{Gray}
    VNT~\cite{pustynnikov2021estimating} & \multicolumn{2}{c|}{SNGP} & - & - & \underline{15.60} & \underline{54.0} & 2.619 \\
    
    ViT~\cite{postnikov2021transformer} & \multicolumn{2}{c|}{$E_{reg}$} & - & - & 15.68 & 53.1 & \underline{2.571} \\  

    \midrule

    \rowcolor{Gray}
    {HiVT$^*$ ($\Sigma=I$)} & $l$GMM & $E_{reg}$ & {\textbf{11.25}} & {\textbf{14.74}} & {\textbf{11.95}} & {\textbf{59.3}} & {\textbf{2.207}} \\  
    
\bottomrule
\end{tabular}
}
\label{table:shifts_challenge}
\end{table}

\subsection{Qualitative Results}
 \cref{fig:qualitative_results} shows qualitative results of our method illustrating the trajectory prediction as well as the OOD score $\hat{\alpha}$ and the uncertainty $\hat{e}$. From left to right, we show two OOD and two ID scenarios. In the OOD scenarios, the trajectory prediction is inaccurate resulting in large prediction errors, as indicated by the blue triangle. The scenarios fall in regions with a low likelihood of the probability density function of the \emph{lGMM}, resulting in high OOD scores. In the ID scenarios, the model can predict a multi-modal future but misses the actual behavior of a lane change in the first scenario and a right turn prediction in the second scenario, which results in high prediction error. The large prediction error is successfully detected by the $E_{reg}$ module, which predicts large uncertainty in both cases. In each situation, despite it being ID or OOD, both modules assess the reliability of the prediction, where \emph{lGMM} detects the OOD scenarios and $E_{reg}$ estimates the uncertainty. 

\subsection{Runtime and Learnable Model Parameters}

\cref{table:runtime} lists the runtime and the number of learnable model parameters of the scene encoder $f_e$, the trajectory prediction decoder $g_p$, the OOD detection $g_{ood}$ and the uncertainty estimation $g_u$.
Assuming that the inferences are executed sequentially, the OOD detection and the uncertainty estimation cause a \SI{0.14}{ms} increase in runtime, which is a relative increase of only $5 \%$.
This low increase in runtime illustrates the benefit of using a joint feature space $h_i$ for all modules.
Previous work mainly relied on ensembles of multiple models (e.g., $K=5$) to achieve reasonable results for OOD detection and uncertainty estimation.
Using $5$ models instead of only $1$ trajectory predictor is a relative runtime increase of $400 \%$.

\begin{table}[t]
\centering
\caption{Evaluation of Runtime and Learnable Model Parameters.}
\resizebox{\linewidth}{!}{
\begin{tabular}{@{}l|c|ccc}
     &  $f_e$ & $g_p$ & $g_{ood}$ & $g_u$ \\ \midrule
     Inference Time & \SI{2}{ms}$^*$ & \SI{0.8}{ms}$^*$ & \SI{0.05}{ms}$^\dag$ & \SI{0.09}{ms}$^*$ \\
     Model Parameters & \SI{1488768}{} & \SI{1203052}{} & \SI{99078}{} & \SI{5712}{} \\
\bottomrule
\end{tabular}
}
\begin{flushleft}
$^*$NVIDIA GeForce RTX 2080 Ti.\quad $^\dag$Intel Core i9-10900X @ 3.7 GHz.
\end{flushleft}
\vspace{-0.5cm}
\label{table:runtime}
\end{table}

\addtolength{\textheight}{-2.6cm}   %

\section{Conclusion}
We presented a trajectory prediction model with joint OOD detection and uncertainty estimation.
The model is composed of the latent Gaussian mixture model and the error regression network to assess the reliability of trajectory prediction in ID as well as in OOD scenarios using a shared scene encoder.
We demonstrated the efficacy of the proposed approach with experimental results on the Shifts dataset.
Our results show that generative modeling of the latent features improves the OOD detection. Additionally, the regression of the prediction error is a simple yet effective way to estimate the current prediction error.
Unlike the prior work, like the ensemble-based methods, our approach can extend existing trajectory prediction models to assess the prediction reliability without retraining the prediction model and with low computational overhead. 
It remains to be investigated, how the OOD score and the uncertainty can be used by a downstream planner.

\section*{Acknowledgment}
The research leading to these results is funded by the German Federal Ministry for Economic Affairs and Energy within the project “KI Delta Learning" (Förderkennzeichen 19A19013A). The authors would like to thank the consortium for the successful cooperation.

\bibliographystyle{iros_style/IEEEtran}
\bibliography{iros_style/IEEEabrv, egbib.bib}

\begin{thebibliography}{10}
\providecommand{\url}[1]{#1}
\csname url@rmstyle\endcsname
\providecommand{\newblock}{\relax}
\providecommand{\bibinfo}[2]{#2}
\providecommand\BIBentrySTDinterwordspacing{\spaceskip=0pt\relax}
\providecommand\BIBentryALTinterwordstretchfactor{4}
\providecommand\BIBentryALTinterwordspacing{\spaceskip=\fontdimen2\font plus
\BIBentryALTinterwordstretchfactor\fontdimen3\font minus
  \fontdimen4\font\relax}
\providecommand\BIBforeignlanguage[2]{{%
\expandafter\ifx\csname l@#1\endcsname\relax
\typeout{** WARNING: IEEEtran.bst: No hyphenation pattern has been}%
\typeout{** loaded for the language `#1'. Using the pattern for}%
\typeout{** the default language instead.}%
\else
\language=\csname l@#1\endcsname
\fi
#2}}

\bibitem{malinin2021shiftsdataset}
A.~Malinin, N.~Band, Y.~Gal, M.~Gales, A.~Ganshin, G.~Chesnokov, A.~Noskov,
  A.~Ploskonosov, L.~Prokhorenkova, I.~Provilkov, V.~Raina, V.~Raina,
  D.~Roginskiy, M.~Shmatova, P.~Tigas, and B.~Yangel, ``Shifts: A dataset of
  real distributional shift across multiple large-scale tasks,'' in
  \emph{Thirty-fifth Conference on Neural Information Processing Systems
  Datasets and Benchmarks Track (Round 2)}, 2021.

\bibitem{Bahari_2022_CVPR}
M.~Bahari, S.~Saadatnejad, A.~Rahimi, M.~Shaverdikondori, A.~H. Shahidzadeh,
  S.-M. Moosavi-Dezfooli, and A.~Alahi, ``Vehicle trajectory prediction works,
  but not everywhere,'' in \emph{Proceedings of the IEEE/CVF Conference on
  Computer Vision and Pattern Recognition (CVPR)}, June 2022, pp.
  17\,123--17\,133.

\bibitem{ivanovic2022propagating}
B.~Ivanovic, Y.~Lin, S.~Shrivastava, P.~Chakravarty, and M.~Pavone,
  ``Propagating state uncertainty through trajectory forecasting,'' in
  \emph{2022 International Conference on Robotics and Automation (ICRA)}.\hskip
  1em plus 0.5em minus 0.4em\relax IEEE, 2022, pp. 2351--2358.

\bibitem{zhou2022hivt}
Z.~Zhou, L.~Ye, J.~Wang, K.~Wu, and K.~Lu, ``Hivt: Hierarchical vector
  transformer for multi-agent motion prediction,'' in \emph{Proceedings of the
  IEEE/CVF Conference on Computer Vision and Pattern Recognition}, 2022, pp.
  8823--8833.

\bibitem{liang2020learning}
M.~Liang, B.~Yang, R.~Hu, Y.~Chen, R.~Liao, S.~Feng, and R.~Urtasun, ``Learning
  lane graph representations for motion forecasting,'' in \emph{Computer
  Vision--ECCV 2020: 16th European Conference, Glasgow, UK, August 23--28,
  2020, Proceedings, Part II 16}.\hskip 1em plus 0.5em minus 0.4em\relax
  Springer, 2020, pp. 541--556.

\bibitem{salzmann2020trajectron++}
T.~Salzmann, B.~Ivanovic, P.~Chakravarty, and M.~Pavone, ``Trajectron++:
  Dynamically-feasible trajectory forecasting with heterogeneous data,'' in
  \emph{European Conference on Computer Vision}.\hskip 1em plus 0.5em minus
  0.4em\relax Springer, 2020, pp. 683--700.

\bibitem{varadarajan2022multipath++}
B.~Varadarajan, A.~Hefny, A.~Srivastava, K.~S. Refaat, N.~Nayakanti,
  A.~Cornman, K.~Chen, B.~Douillard, C.~P. Lam, D.~Anguelov, \emph{et~al.},
  ``Multipath++: Efficient information fusion and trajectory aggregation for
  behavior prediction,'' in \emph{2022 International Conference on Robotics and
  Automation (ICRA)}.\hskip 1em plus 0.5em minus 0.4em\relax IEEE, 2022, pp.
  7814--7821.

\bibitem{wiederer2022abenchmark}
J.~Wiederer, J.~Schmidt, U.~Kressel, K.~Dietmayer, and V.~Belagiannis, ``A
  benchmark for unsupervised anomaly detection in multi-agent trajectories,''
  in \emph{2022 IEEE 25th International Conference on Intelligent
  Transportation Systems (ITSC)}, 2022, pp. 130--137.

\bibitem{chakraborty2023structural}
N.~Chakraborty, A.~Hasan, S.~Liu, T.~Ji, W.~Liang, D.~L. McPherson, and
  K.~Driggs-Campbell, ``Structural attention-based recurrent variational
  autoencoder for highway vehicle anomaly detection,'' \emph{arXiv preprint
  arXiv:2301.03634}, 2023.

\bibitem{pustynnikov2021estimating}
A.~Pustynnikov and D.~Eremeev, ``Estimating uncertainty for vehicle motion
  prediction on yandex shifts dataset,'' \emph{arXiv preprint
  arXiv:2112.08355}, 2021.

\bibitem{postnikov2021transformer}
A.~Postnikov, A.~Gamayunov, and G.~Ferrer, ``Transformer based trajectory
  prediction,'' \emph{arXiv preprint arXiv:2112.04350}, 2021.

\bibitem{Gilles2022}
T.~Gilles, S.~Sabatini, D.~V. Tsishkou, B.~Stanciulescu, and F.~Moutarde,
  ``Uncertainty estimation for cross-dataset performance in trajectory
  prediction,'' \emph{2022 International Conference on Robotics and Automation
  Workshop (ICRA Workshop)}, 2022.

\bibitem{lakshminarayanan2017simple}
B.~Lakshminarayanan, A.~Pritzel, and C.~Blundell, ``Simple and scalable
  predictive uncertainty estimation using deep ensembles,'' \emph{Advances in
  neural information processing systems}, vol.~30, 2017.

\bibitem{gal2016dropout}
Y.~Gal and Z.~Ghahramani, ``Dropout as a bayesian approximation: Representing
  model uncertainty in deep learning,'' in \emph{international conference on
  machine learning}.\hskip 1em plus 0.5em minus 0.4em\relax PMLR, 2016, pp.
  1050--1059.

\bibitem{lee2018simple}
K.~Lee, K.~Lee, H.~Lee, and J.~Shin, ``A simple unified framework for detecting
  out-of-distribution samples and adversarial attacks,'' \emph{Advances in
  neural information processing systems}, vol.~31, 2018.

\bibitem{Ahuja2019probabilistic}
N.~A. Ahuja, I.~Ndiour, T.~Kalyanpur, and O.~Tickoo, ``Probabilistic modeling
  of deep features for out-of-distribution and adversarial detection,''
  \emph{NeurIPS Bayesian Deep Learning Workshop}, 2019.

\bibitem{Chai2019MultiPathMP}
Y.~Chai, B.~Sapp, M.~Bansal, and D.~Anguelov, ``Multipath: Multiple
  probabilistic anchor trajectory hypotheses for behavior prediction,'' in
  \emph{CoRL}, 2019.

\bibitem{strohbeck2020multiple}
J.~Strohbeck, V.~Belagiannis, J.~Müller, M.~Schreiber, M.~Herrmann, D.~Wolf,
  and M.~Buchholz, ``Multiple trajectory prediction with deep temporal and
  spatial convolutional neural networks,'' in \emph{2020 IEEE/RSJ International
  Conference on Intelligent Robots and Systems (IROS)}, 2020, pp. 1992--1998.

\bibitem{gao2020vectornet}
J.~Gao, C.~Sun, H.~Zhao, Y.~Shen, D.~Anguelov, C.~Li, and C.~Schmid,
  ``Vectornet: Encoding hd maps and agent dynamics from vectorized
  representation,'' in \emph{Proceedings of the IEEE/CVF Conference on Computer
  Vision and Pattern Recognition}, 2020, pp. 11\,525--11\,533.

\bibitem{schmidt2022crat}
J.~Schmidt, J.~Jordan, F.~Gritschneder, and K.~Dietmayer, ``Crat-pred: Vehicle
  trajectory prediction with crystal graph convolutional neural networks and
  multi-head self-attention,'' in \emph{2022 International Conference on
  Robotics and Automation (ICRA)}.\hskip 1em plus 0.5em minus 0.4em\relax IEEE
  Press, 2022, p. 7799–7805.

\bibitem{chang2019argoverse}
M.-F. Chang, J.~Lambert, P.~Sangkloy, J.~Singh, S.~Bak, A.~Hartnett, D.~Wang,
  P.~Carr, S.~Lucey, D.~Ramanan, \emph{et~al.}, ``Argoverse: 3d tracking and
  forecasting with rich maps,'' in \emph{Proceedings of the IEEE/CVF Conference
  on Computer Vision and Pattern Recognition}, 2019, pp. 8748--8757.

\bibitem{hsu2020generalized}
Y.-C. Hsu, Y.~Shen, H.~Jin, and Z.~Kira, ``Generalized odin: Detecting
  out-of-distribution image without learning from out-of-distribution data,''
  in \emph{Proceedings of the IEEE/CVF Conference on Computer Vision and
  Pattern Recognition}, 2020, pp. 10\,951--10\,960.

\bibitem{hornauer2023heatmap}
J.~Hornauer and V.~Belagiannis, ``Heatmap-based out-of-distribution
  detection,'' in \emph{Proceedings of the IEEE/CVF Winter Conference on
  Applications of Computer Vision}, 2023, pp. 2603--2612.

\bibitem{ren2019likelihood}
J.~Ren, P.~J. Liu, E.~Fertig, J.~Snoek, R.~Poplin, M.~Depristo, J.~Dillon, and
  B.~Lakshminarayanan, ``Likelihood ratios for out-of-distribution detection,''
  \emph{Advances in neural information processing systems}, vol.~32, 2019.

\bibitem{Richter2017SafeVN}
C.~Richter and N.~Roy, ``Safe visual navigation via deep learning and novelty
  detection,'' in \emph{Robotics: Science and Systems}, 2017.

\bibitem{sindhwani2020unsupervised}
V.~Sindhwani, H.~Sidahmed, K.~Choromanski, and B.~Jones, ``Unsupervised anomaly
  detection for self-flying delivery drones,'' in \emph{2020 IEEE International
  Conference on Robotics and Automation (ICRA)}.\hskip 1em plus 0.5em minus
  0.4em\relax IEEE, 2020, pp. 186--192.

\bibitem{wiederer22anomaly}
J.~Wiederer, A.~Bouazizi, M.~Troina, U.~Kressel, and V.~Belagiannis, ``Anomaly
  detection in multi-agent trajectories for automated driving,'' in
  \emph{Proceedings of the 5th Conference on Robot Learning}, ser. Proceedings
  of Machine Learning Research, A.~Faust, D.~Hsu, and G.~Neumann, Eds., vol.
  164.\hskip 1em plus 0.5em minus 0.4em\relax PMLR, 08--11 Nov 2022, pp.
  1223--1233.

\bibitem{schmidt2023scene}
T.~Monninger, J.~Schmidt, J.~Rupprecht, D.~Raba, J.~Jordan, D.~Frank, S.~Staab,
  and K.~Dietmayer, ``Scene: Reasoning about traffic scenes using heterogeneous
  graph neural networks,'' \emph{IEEE Robotics and Automation Letters}, vol.~8,
  no.~3, pp. 1531--1538, 2023.

\bibitem{bishop1994mixture}
C.~M. Bishop, ``Mixture density networks,'' \emph{Aston University}, 1994.

\bibitem{bishop2006pattern}
C.~M. Bishop and N.~M. Nasrabadi, \emph{Pattern recognition and machine
  learning}.\hskip 1em plus 0.5em minus 0.4em\relax Springer, 2006, vol.~4,
  no.~4.

\bibitem{ivanovic2019trajectron}
B.~Ivanovic and M.~Pavone, ``The trajectron: Probabilistic multi-agent
  trajectory modeling with dynamic spatiotemporal graphs,'' in
  \emph{Proceedings of the IEEE/CVF International Conference on Computer
  Vision}, 2019, pp. 2375--2384.

\bibitem{DBLP:journals/corr/HendrycksG16c}
D.~Hendrycks and K.~Gimpel, ``A baseline for detecting misclassified and
  out-of-distribution examples in neural networks,'' \emph{CoRR}, vol.
  abs/1610.02136, 2016.

\bibitem{Loshchilov2019DecoupledWD}
I.~Loshchilov and F.~Hutter, ``Decoupled weight decay regularization,'' in
  \emph{ICLR}, 2019.

\bibitem{Loshchilov2017SGDR}
I.~Loshchilov and F.~{Hutter}, ``{SGDR}: Stochastic gradient descent with warm
  restarts,'' in \emph{International Conference on Learning Representations},
  2017.

\end{thebibliography}

\end{document}